%% file: main.tex
\documentclass[letterpaper, 10 pt, conference]{ieeeconf}  % Comment this line out if you need a4paper

\IEEEoverridecommandlockouts                              % This command is only needed if 
\usepackage{graphicx}

\usepackage{amsmath} % assumes amsmath package installed
\usepackage{amssymb}  % assumes amsmath package installed
\usepackage{hyperref} % Required for hyperlinks
\usepackage[
backend=biber,
style=numeric,
sorting=none
]{biblatex}
\usepackage[table]{xcolor}
\usepackage{amsmath}

\input{definitions}

\title{\LARGE \bf
Looking Inside Out: Anticipating Driver Intent From Videos
}

\author{
Yung-Chi Kung$^{*}$ and Arthur Zhang$^{*}$ and Junmin Wang and Joydeep Biswas% <-this % stops a space
\thanks{*Equal Contribution.}% <-this % stops a space
\thanks{Y. Kung, A. Zhang, J. Wang, and J. Biswas are with the University of Texas at Austin, Texas, USA.
\tt\small{\{yung-chi.kung\}@utexas.edu \tt\small\{jmwang\}@austin.utexas.edu \tt\small\{arthurz, joydeepb\}@cs.utexas.edu}
}
}

\begin{document}

\maketitle
\thispagestyle{empty}
\pagestyle{empty}

%%%%%%%%%%%%%%%%%%%%%%%%%%%%%%%%%%%%%%%%%%%%%%%%%%%%%%%%%%%%%%%%%%%%%%%%%%%%%%%%
\begin{abstract}

Anticipating driver intention is an important task when vehicles of mixed and varying levels of human/machine autonomy share roadways. Driver intention can be leveraged to improve road safety, such as warning surrounding vehicles in the event the driver is attempting a dangerous maneuver. In this work, we propose a novel method of utilizing in-cabin and external camera data to improve state-of-the-art (SOTA) performance in predicting future driver actions. Compared to existing methods, our approach explicitly extracts object and road-level features from external camera data, which we demonstrate are important features for predicting driver intention. Using our handcrafted features as inputs for both a transformer and an LSTM-based architecture, we empirically show that jointly utilizing in-cabin and external features improves performance compared to using in-cabin features alone. Furthermore, our models predict driver maneuvers more accurately and earlier than existing approaches, with an accuracy of 87.5\% and an average prediction time of 4.35 seconds before the maneuver takes place. We release our model configurations and training scripts on \url{https://github.com/ykung83/Driver-Intent-Prediction}.
\end{abstract}

%%%%%%%%%%%%%%%%%%%%%%%%%%%%%%%%%%%%%%%%%%%%%%%%%%%%%%%%%%%%%%%%%%%%%%%%%%%%%%%%

\input{introduction}

\input{relatedworks}

\input{methodology}

\input{experiment}

\input{results}

\input{conclusion}  
% \section{INTRODUCTION}

%%%%%%%%%%%%%%%%%%%%%%%%%%%%%%%%%%%%%%%%%%%%%%%%%%%%%%%%%%%%%%%%%%%%%%%%%%%%%%%%

\printbibliography[]
% \bibliographystyle{IEEEtran}
% \bibliography{refs}

\end{document}

%% file: definitions.tex
\usepackage{xcolor}

\newcommand{\figlabel}[1]{\label{fig:#1}}
\newcommand{\figref}[1]{Fig.~\ref{fig:#1}}

%% file: introduction.tex
\section{Introduction}

The number of vehicles being driven is continuously increasing, but less than half of all drivers follow even basic safety conduct like turning on a blinker before performing a lane change~\cite{srefstats1}. To improve road safety, many safety-centric Advanced Driver Assistance Systems (ADAS) and Automated Driving Systems (ADS)~\cite{srefadas1, srefadas2} have been designed to anticipate the actions of the driver and provide warnings or assistive actions. These approaches measure success using the prediction accuracy and average prediction time before the maneuver takes place (time-until-maneuver, TUM).

To predict driver intentions, both in-cabin and external information should be utilized. It is well documented that cephalo-ocular cues are an excellent indicator of driver intent~\cite{srefgaze1,srefgaze2}. However, the use of external data has been shown~\cite{mref1} to decrease accuracy and TUM. Consequently, follow-up work~\cite{mref3} purposely choose not to utilize external sensing, relying on internal camera and vehicle dynamics data. While there exist methods focused on the fusion of external data streams with internal data~\cite{srefstats3}, they cannot match the SOTA performance.

\begin{figure}[ht]
    \centering
    \includegraphics[width=\linewidth]{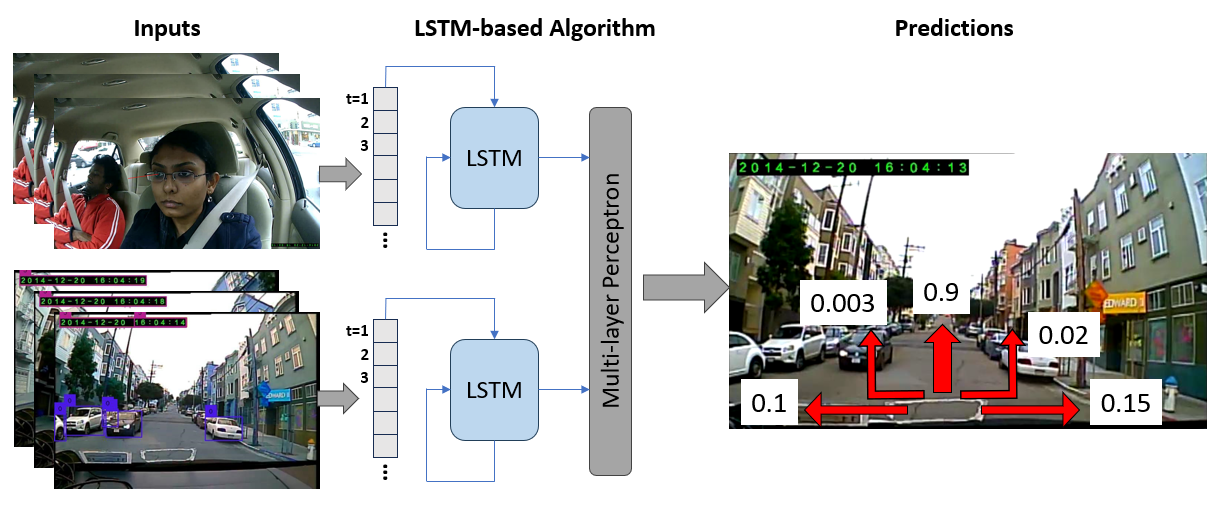}
    \includegraphics[width=\linewidth]{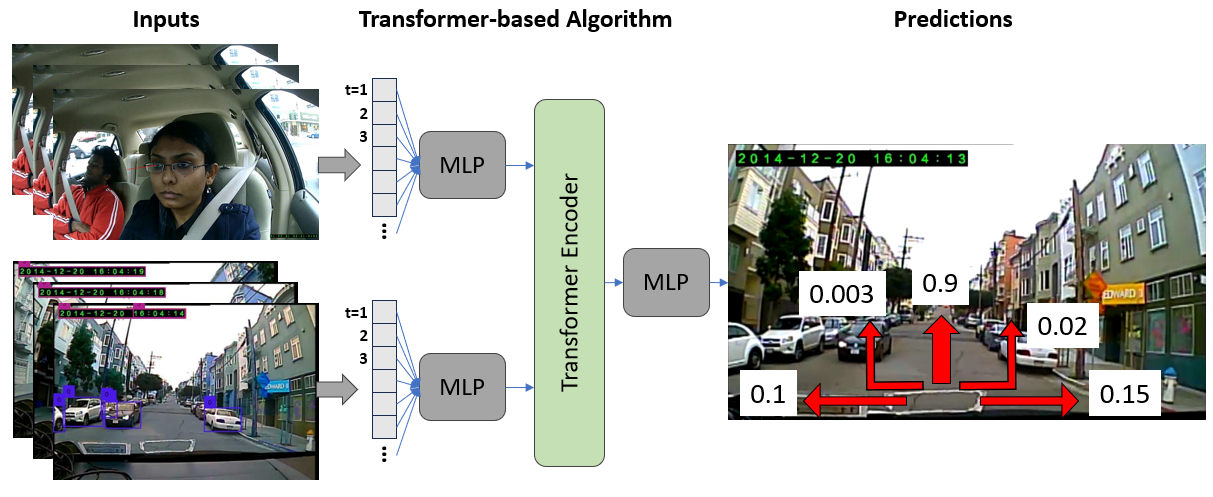}
    \caption{\textbf{Predicting Driver Intent} Our proposed LSTM and transformer-based architectures use eye gaze, head pose, object detections, and road features from a real-world driving dataset to outperform state-of-the-art methods for driver intent prediction. }
    \label{fig:cover}
\end{figure}

Despite these findings, we hypothesize that external sensing provides invaluable information for understanding driver intent. Vehicle surroundings provide context that may explain observed cephalo-ocular cues and communicate which maneuvers are possible. We propose explicitly extracting object and road level features from road camera videos instead of learning these features in an end-to-end manner. We employ two model architectures proficient in handling time dependencies to combine feature vectors from both in-cabin and external cameras. This approach surpasses all other methods when evaluated on a real-world dataset for predicting driver intentions~\cite{brain4cars}. A high-level overview of our proposed method is shown in Figure \ref{fig:cover}. In the remainder of the manuscript, we review prior work, describe the experiment setup, explain our model architectures, and conclude with an analysis of our results.

%% file: relatedworks.tex
\section{Related Work}
A considerable amount of work has been done on driver maneuver prediction. Early efforts~\cite{sref3man1, sref3man2} predict three different driving maneuvers: straight-line driving, left lane change, and right lane change. Later attempts \cite{mref1, mref2, mref3} expand the action space to include left turn and right turn. We analyze both 3- and 5-maneuver prediction methods but focus on 5-maneuver prediction as it more closely resembles the true action space available to drivers.

% Despite evaluating our method on 5-maneuver prediction, We will focus on works involving 5-maneuver prediction as they more closely resemble the true action space available to drivers.
% Additionally, we place that use other modalities like vehicle dynamics~\cite{srefsteer1} or gaze~\cite{srefgaze1, srefgaze2} to predict driver intent.

Z. Hao et al~\cite{sref3man2} use a gated-recurrent unit (GRU) with an attention mechanism for the 3-driving maneuver problem. Their model uses solely vehicle dynamics to achieve high accuracy and precision when predicting one second before the driving maneuver takes place. However, both accuracy and precision decrease substantially if asked to predict at an earlier time step. N. Zhao et al~\cite{sref3man1} employ a Convolutional Neural Network (CNN) and a Long-Short-Term-Memory (LSTM) based network to interpret driving dynamics and roadway information provided by the simulator for 3-maneuver prediction. Their method gives good accuracy but does not provide information on how early the network can predict driver intent before it occurs and only evaluates on simulated driving data. 

P. Gebert et al~\cite{mref1} utilize an end-to-end CNN and LSTM-based network for the 5-maneuver problem. Their approach feeds the raw interior camera data through an optical flow estimation algorithm. They provide this output to a CNN for classification and feature extraction and use these features in an LSTM for prediction. Their method has an accuracy of 83.12\% and an average prediction time of 4.07 seconds using only interior camera data. When using external video data, the accuracy drops to 75.5\%. The paper~\cite{mref1} also released their real-world driving dataset, which we use as a benchmark. The other SOTA method is from N. Khairdoost et al~\cite{mref3}. They generate an LSTM-based network with driver gaze, head pose, and vehicle dynamics data as inputs. The work expresses gaze information as a histogram. Each bin in the histogram correlates with a region of space in the driver's field of view. Despite video footage of the exterior being available, the authors choose not to use it in their network. Their method has an accuracy of 84.2\% and an average prediction time of 3.6 seconds. 

The uni-modal study by L. Li and P. Li~\cite{srefsteer1} shows that there are only significant correlations between vehicle dynamics and driving maneuvers 0.55 seconds before the maneuver takes place. This means that when predicting driver intent for longer TUM scenarios, vehicle dynamics will not play a significant role. M. Hofbauer et al \cite{srefgaze1} found that regions of interest and situational awareness can be predicted from driver gaze. This may allow us to better understand the driver's priorities and infer the intended maneuver well before it is executed. A. Kar~\cite{srefgaze2} provides a detailed list of different gaze types and their intent. Some examples include fixation, where the eye moves at less than 100 degrees per millisecond and is indicative of cognitive processing and attention; saccade, where the eye moves between 100 and 700 degrees per second and is indicative of moving between targets of interest; and smooth pursuit, where the eye moves at less than 100 degrees per second based on the targets speed and is indicative of target tracking. This suggests that a high-precision list of gaze locations ordered in time would provide valuable insights into the driver's intentions.

%% file: methodology.tex
\section{methodology}

We aim to increase the driver intent prediction accuracy and extend the average TUM in this work. Since vehicle dynamics are only useful for short term maneuver prediction~\cite{srefsteer1}, it is not used. Instead, we extract handcrafted features from the in-cabin and external camera data. We follow the evaluation procedure used by current state-of-the-art (SOTA) methods to ensure a fair comparison of results.

\subsection{Data}
We train and evaluate our methods on the publicly available Brains4Cars~\cite{brain4cars} dataset, which contains a collection of naturalistic driving maneuvers for driver action prediction containing RGB videos of both vehicle cabin and external road view. While the original dataset reports 700 vehicle maneuver videos, a portion of the training data is missing and 634 videos are publicly available. These videos are comprised of 234 driving straight, 124 left lane change, 58 left turn, 123 right lane change, and 55 right turn maneuvers. Each video is 5 seconds and 150 frames long. Per \cite{mref1}, these five-second snippets are from 6 seconds to 1 second before the maneuver. Following standard convention, the time of maneuver is based on the time the vehicle crosses the lane line. Using this dataset allows direct result comparisons between our methods and the prior work \cite{mref1, brain4cars,brain4cars2} because they are all based on the same data. 

\subsection{Interior Camera}
Much like Leonhardt et al~\cite{mref3}, we extract gaze and head pose information from the interior camera. The difference is that in\cite{mref3}, they have a built-in non-contact 3D gaze and head pose tracker running at 60 Hz while the dataset from Brains4Cars only provides an RGB video feed from the interior camera at 30 Hz. \cite{srefgaze2} provides some references to extract eye gaze from a single stationary video. 

We use MediaPipe to extract face landmarks from the driver in 2D and define a list of generic face landmarks with their coordinates in 3D. Using the solvePnP solver from OpenCV, we use these two lists of landmarks to estimate the projection of the rotation and translation of the driver's face onto a 2D plane using Eq. \ref{eq:gaze}. $S$ is an unknown scale factor, $u$ and $v$ are points from the 2D image, $M$ is an estimate of the camera matrix, $R$ is the rotation matrix, $x$,$y$, and $z$ are from the tuned 3D model, and $t$ is the translation vector.

\begin{equation}
    S \begin{bmatrix}
        u \\ v \\ 1 
    \end{bmatrix} = M ( R \begin{bmatrix}
        x \\ y \\ z
    \end{bmatrix} + t ) \label{eq:gaze}
\end{equation}

\begin{figure}[hbp]
    \centering
    \includegraphics[width=\linewidth]{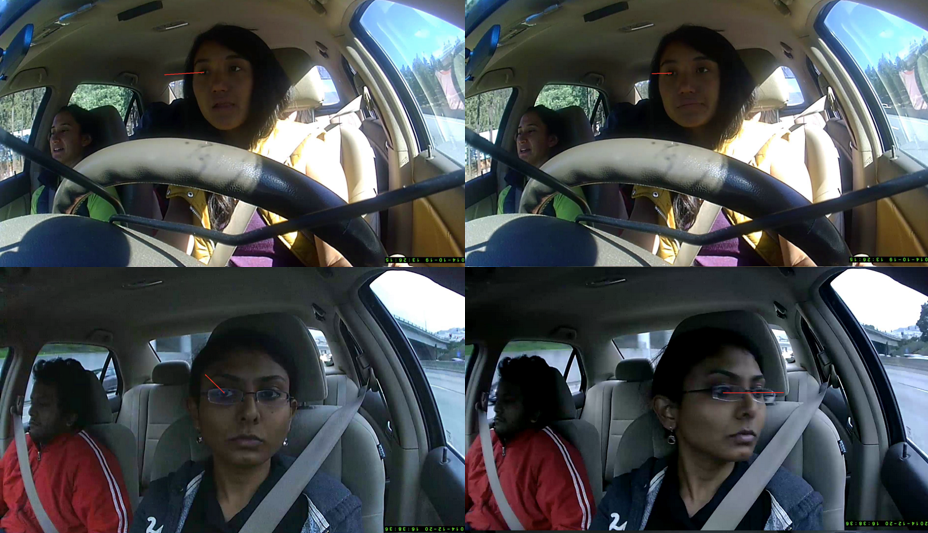}
    \caption{Qualitative results of the gaze preprocessing algorithm on the Brains4Cars dataset. The red line projects their gaze to a point on the imaginary plane.}
    \label{fig:GPP}
\end{figure}

The rotation and translation vectors give us information on the direction the driver is facing. This is our approximation for the driver's head pose. Next, we get the 3D coordinates of the pupils by using the estimateAffline3D function from OpenCV to estimate the transformation between the estimated face coordinates and the model of a generic face. This gives us a 3D representation of the driver's pupils. Once we have the eye center and the pupil location, we can project a line through those two points onto the same 2D plane used for the head pose to get the location of the gaze. This is illustrated in Figure \ref{fig:GPP}.

The representation of the driver's head pose and gaze is a 4-dimensional vector containing the x and y coordinates of the intersection of the projected line from the driver's face and eyes with an imaginary 2D plane representing the windshield of the car. We choose this representation of the driver's gaze because it provides sufficient information for data-driven algorithms to distinguish fine-grained eye movements that provide different connotations regarding driver intent~\cite{srefgaze2}. Prior histogram approaches~\cite{mref3} would not be precise enough to make this differentiation. 

\subsection{Exterior Camera}

Our method uses the exterior camera to add lane and object-level information inputs to our model. We are the first method that directly incorporates object-level information into our model. Comprehending the context of objects allows methods to determine if maneuvers are unsafe due to nearby objects. We leverage Grounding Dino~\cite{liu2023grounding}, a SOTA zero-shot 2D object detector to detect the following object classes: \textsc{Car}, \textsc{Bicycle}, \textsc{Person}, \textsc{Traffic Sign}, \textsc{Traffic Light}, and \textsc{Date}. We omit the \textsc{Date} class from the model and store the bounding box centers, height, width, and class ID as model inputs. \figref{roadqual} shows sample object detections on the Brains4Cars dataset. For computational reasons, we provide only the top 5 largest bounding boxes by area for each frame to the model. 

 For lane information, we use the ground truth lane labels in Brains4Cars, which contain the vehicle lane position, number of lanes, and whether the car is near a road intersection. This is useful because if there is no lane to the left of the driver, it should be a significant indicator that the driver is not attempting a left lane change. 

We use a 28-dimensional vector representation for the exterior camera, comprised of a 25-dimensional vector that describes the locations of surrounding objects and a 3-dimensional vector that represents the composition of lanes around the vehicle. 

\subsection{Evaluation}
Like other SOTA methods, our method is evaluated on model accuracy, F1 score, and the average TUM that the model can correctly predict the driver's intentions. The accuracy and F1 scores are based on the performance of the models when trained and tested on the full 5 seconds of driving. We refer to this as Zero-time-to-maneuver. The average TUM is assessed by training the model on various time increments (1, 2, 3, 4, and 5 seconds) of driving data and computing the prediction accuracy at the corresponding time intervals before the maneuver actually occurs. We refer to this as Varying-time-to-maneuver. This is consistent with the approach taken by \cite{mref1}. We train and evaluate with ten-fold cross-validation and report the average performance across all splits for each method.

\begin{figure}[hbp]
  \centering
  \includegraphics[width=\linewidth]{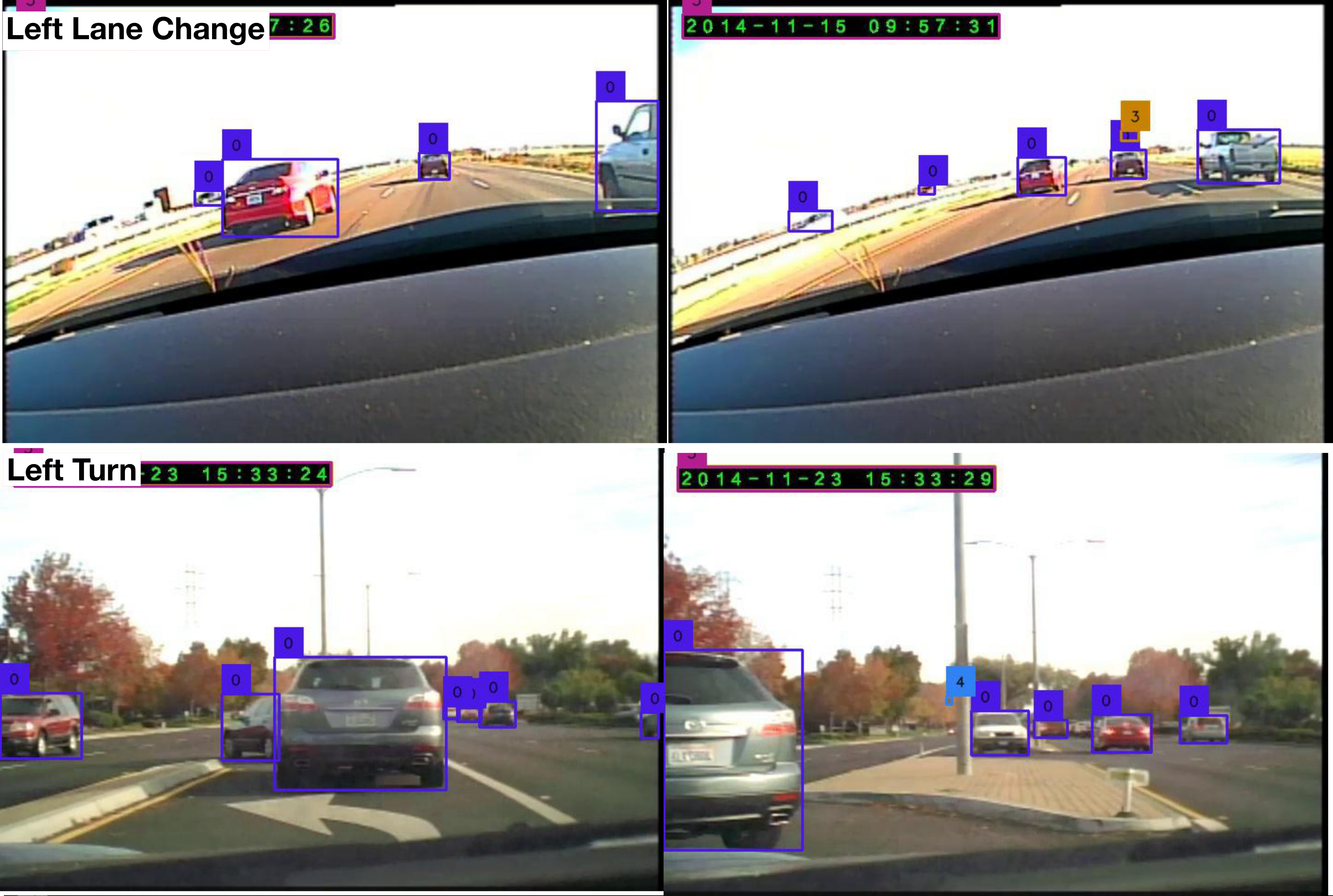}
  \caption{Object detection pre-processing results on the Brains4Cars dataset~\cite{brain4cars}. The object classes Car, Bicycle, Person, Traffic Sign, Traffic Light, and Date are mapped to classes 0, 1, 2, 3, 4, and 5 in the images shown.}
  \figlabel{roadqual}
\end{figure}

%% file: experiment.tex
\section{Experiments}

We propose two machine learning algorithms to predict the driver's intent. The first is based on fusing multiple long-short-term-memory (LSTM) units which we refer to as the F-LSTM. The LSTM is a standard method of predicting driver intent and is used by both SOTA methods that our algorithms are evaluated against \cite{mref3} \cite{mref1}. This makes it a good baseline to compare with the SOTA to evaluate if the hand-tuned features we use as inputs improve prediction accuracy. The second algorithm fuses multiple streams of data using a transformer architecture which we refer to as F-TF. This method will be evaluated against the F-TF to see if it can learn long-term dependencies that would be difficult to capture with an LSTM-based algorithm.

\subsection{F-LSTM}
The LSTM-based architecture is a natural choice given that the problem is inherently dependent on time. LSTMs store and interpret data as a hidden vector, and propagate this hidden vector along each time step. This characteristic is desirable because the inputs are of different modalities and this hidden vector may be used to project their qualities into a common representation. 

\begin{figure}[htbp]
    \centering
`   \includegraphics[width=\linewidth]{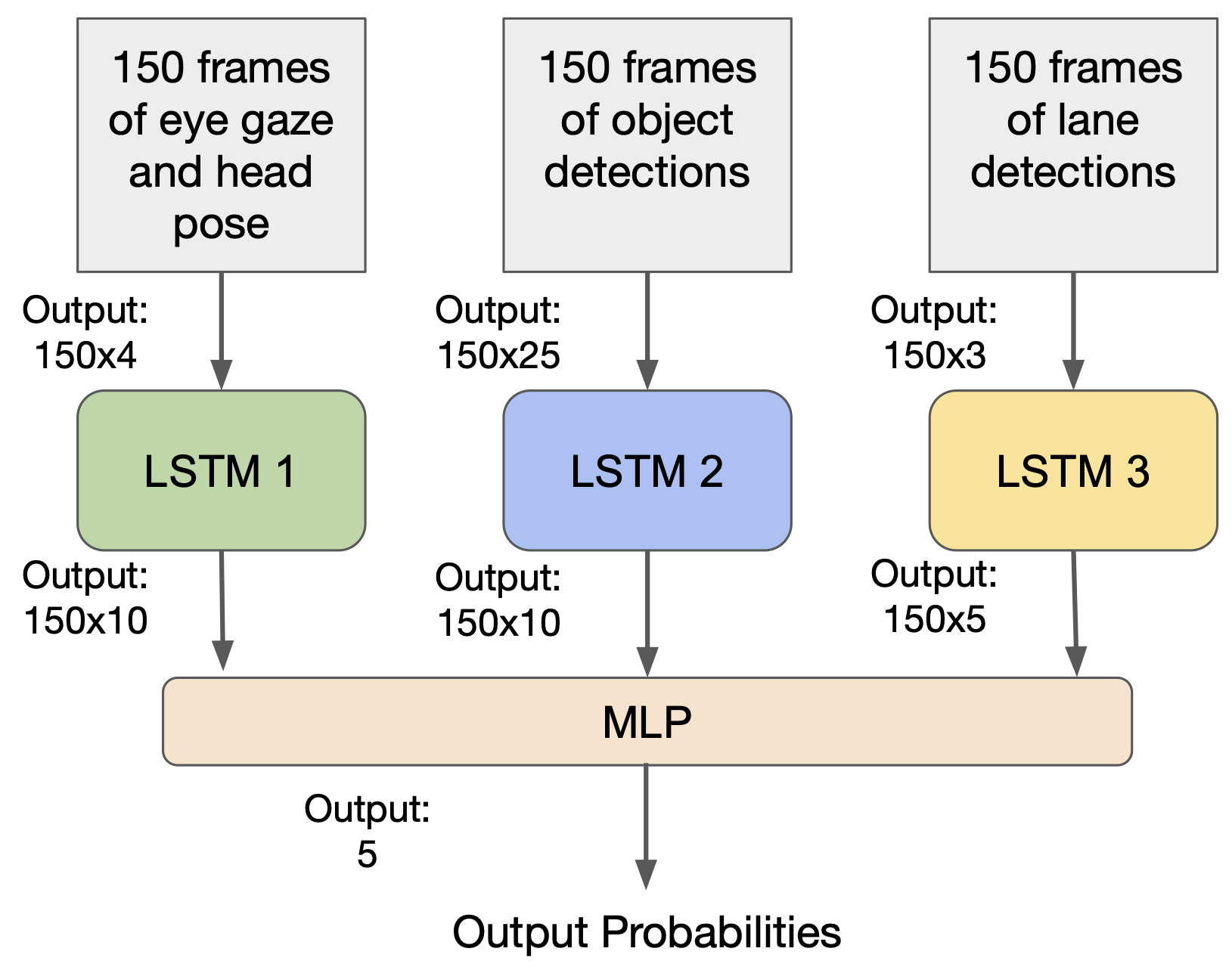}
    \caption{LSTM-based architecture. The MLP is composed of a linear layer with ReLU activation followed by a linear layer with sigmoid activation. }
    \label{fig:FLSTM}
\end{figure}

Figure \ref{fig:FLSTM} shows the architecture of the F-LSTM. Separate LSTMs are used to accept the inputs for head pose and gaze, vehicle objects, and lane detections. This architecture allows each LSTM to specialize in interpreting separate modes of data. LSTM 1 accepts gaze and head pose information and has a hidden dimension of 10. LSTM 2 accepts lane information and has a hidden dimension of 5. LSTM 3 accepts surrounding vehicle information and has a hidden dimension of 10. The outputs of all three LSTMs are then flattened and fully connected to a multi-layer perception (MLP). The MLP consists of a 100 dimension fully connected layer with ReLU activation followed by a 5 dimension fully connected layer with sigmoid activation. The output from the last fully connected layer is used to predict driver actions. Cross-entropy loss is used to train the model.

The same architecture is also used for the time-varying version of the problem. The only difference is that the training sequences are no longer the full 150 frames but a combination of the first 30, 60, 90, 120, and 150 frames of the original data. The extra spaces are padded with zeros. This keeps the time between predictions consistent with \cite{mref1} to allow for a fair comparison. 

\subsection{F-TF}

\begin{figure}[htbp]
    \centering
    \includegraphics[width=\linewidth]{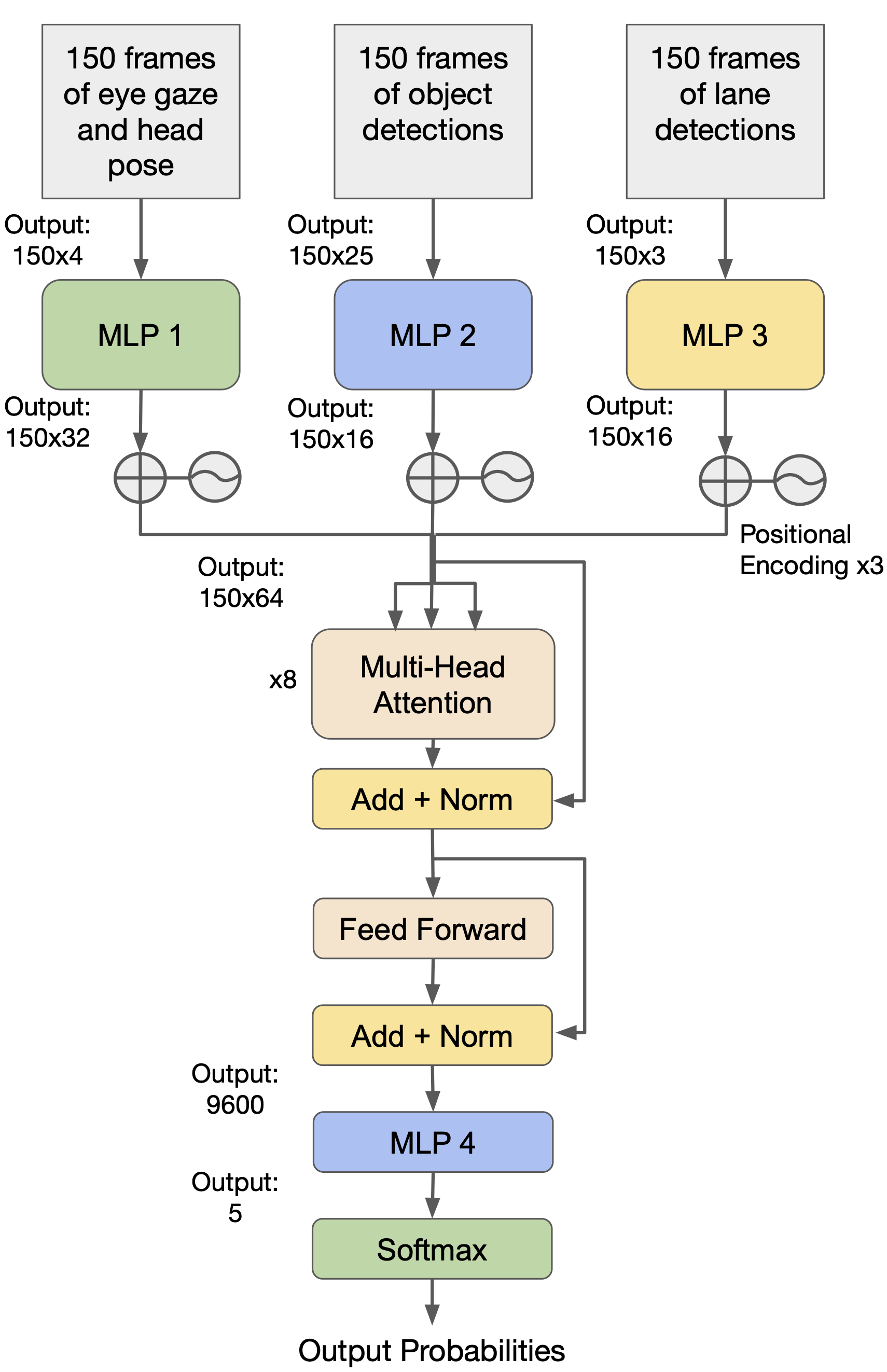}
    \caption{Transformer-based architecture. All MLPs are composed of a Linear, ReLU, and Linear layer. }
    \label{fig:transformerarchitecture}
\end{figure}

While LSTM-based architectures have proven successful for sequential tasks, prior works~\cite{informatik2003} demonstrate that they are adversely affected by long-range time dependencies due to increasing path length for signals. However, the self-attention module in transformer architectures reduces the path length, which can be leveraged to learn long-range correlations in sequential tasks.

Figure~\ref{fig:transformerarchitecture} describes our transformer architecture. We use three input modalities: object detections, road/intersection data, and driver gaze with head pose. This gives feature vectors $\mathbb{R}^{B \times F \times P}$, where B is the batch size, F is the number of images in the sequence, and P is the number of dimensions in each feature vector. \figref{transformerarchitecture} states the size of P for each input modality: 4 features for gaze and head pose; 25 features for object detections; and 3 features for lane information. Each processed vector is linearly projected to a common representation and added with a 1D sinusoidal positional embedding before they are all concatenated to form a unified latent vector. The positional embedding is varied with the time dimension to retain temporal information. We use separate trainable linear projections for each processed vector because each vector is different in both scale and resolution. Similar to the F-LSTM architecture, we find that using a single linear projection results in worse performance than using three separate trainable linear projections. The MLP output vector sizes are 32, 16, and 16 for the in-cabin, object, and road, which provides the same representational power between the in-cabin and external information. 

We perform self-attention between all latent vectors for a single image sequence before feeding them to a standard feedforward module~\cite{VaswaniAttention2017}. This allows our model to represent long and short-term time dependencies with the same path length. Finally, we flatten the feedforward output to a 9600 dimensional vector and project this with a classification head, implemented as an MLP with one hidden layer. The MLP outputs a 5-dimensional feature vector to a softmax function that represents the driver intent probability vector. For the zero-time-to-maneuver experiments, we train using the full 150 frames of data. In the time-varying benchmark, we follow the same training setup as the F-LSTM for consistency.

\subsection{Ablative Testing}
Additional ablative testing is conducted on the F-LSTM and F-TF models to measure the performance contribution of the exterior camera's handcrafted features. The ablative tests for the F-LSTM and the F-TF are named F-LSTM-A and F-TF-A respectively. These tests compare the performance of our F-LSTM and F-TF models with and without the external camera features. The original models are trained and tested using both modes of data while the ablative models are trained and tested using only features from the interior camera. The results are shown in Table \ref{tab:subtablelstm}.

\begin{figure}[htbp]
    \centering
    \includegraphics[width=\linewidth]{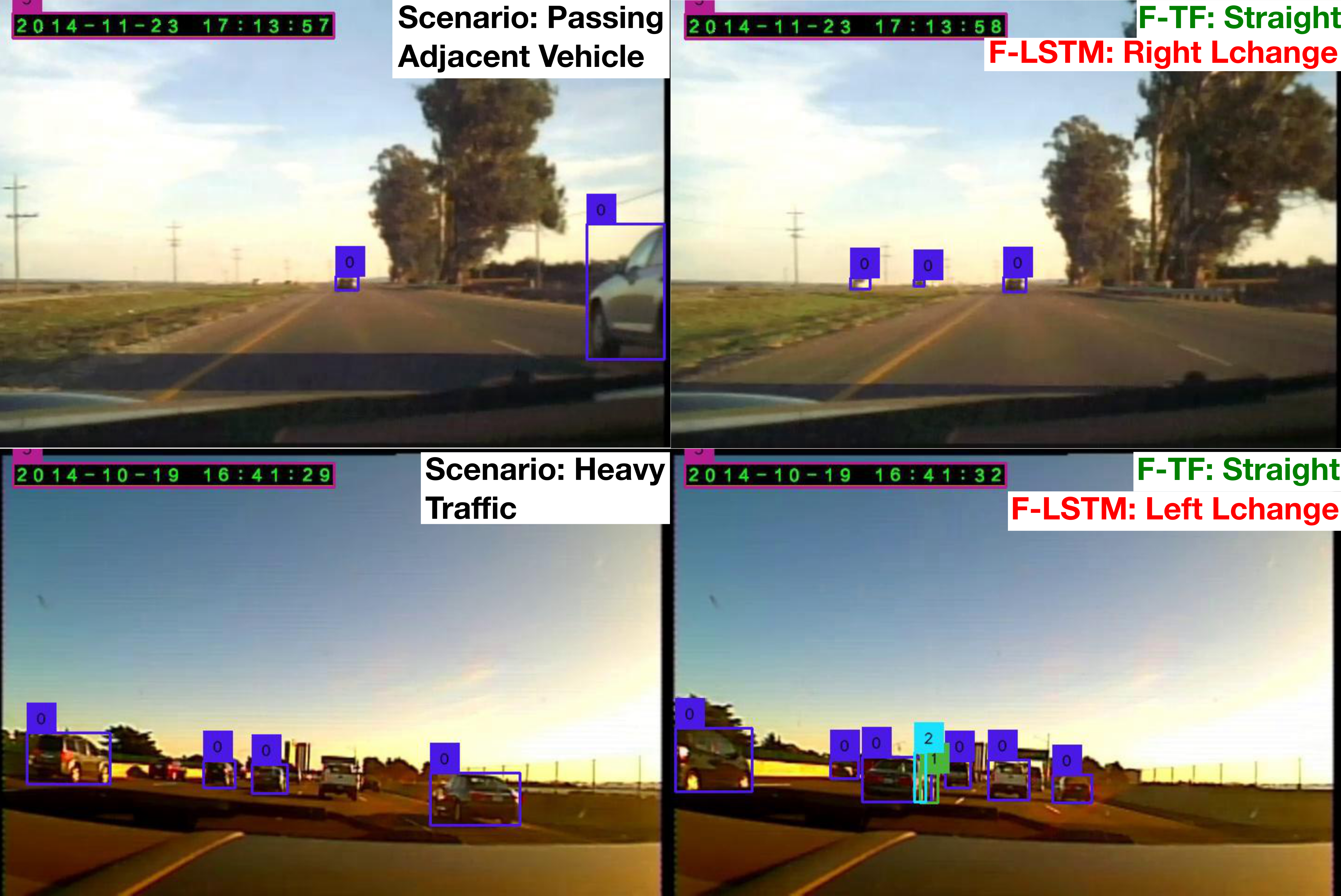}
    \caption{Driver intent prediction scenarios that require long-term dependency understanding. Our transformer and LSTM based architecture are abbreviated as F-TF and F-LSTM. Green and red text indicate correct and incorrect predictions respectively.}
    \label{fig:transformervslstm}
\end{figure}

\begin{table}[htbp]
\caption{\label{tab:subtablelstm} Ablative experiments on the efficacy of the handcrafted features of the exterior.}
\begin{center}
\renewcommand{\arraystretch}{1.2}
\begin{tabular}{p{2.1cm} c c c c }
\hline
Driver Maneuver &  F-LSTM-A &  F-LSTM &  F-TF-A & F-TF \\
\hline
 Straight driving  & 67.3 & 87.0 &  68.7 & 90.9   \\
 Left turn  & 69.2 & 86.8 &  81.0 & 85.1   \\
 Left lane change  & 64.3 & 91.8 &  77.9 & 91.4   \\
 Right turn  & 63.2 & 85.6 &  86.0 & 83.8   \\
 Right lane change  & 65.5 & 84.0 &  79.0 & 86.3   \\
 \hline
 Overall accuracy & 66.2 & 87.2 & 78.1 & 87.5 \\
\hline
\end{tabular}
\renewcommand{\arraystretch}{1}
\end{center}
\end{table}

Table~\ref{tab:subtablelstm} demonstrates that the F-LSTM and F-TF significantly outperform their counterparts that were only trained with in-cabin features. This reinforces the idea that the hand-crafted feature space we designed to describe the exterior view is helpful for predicting driver intent. 

%% file: results.tex
\section{results}

Table~\ref{tab:maintable} compared our methods' results against similar works. We observe that both the F-LSTM and F-TF algorithms outperform other SOTA methods by a significant margin.

\subsection{Zero-time-to-maneuver}

The F-LSTM can be directly compared with the methods from \cite{mref1} and \cite{mref3} since they all use an LSTM-based architecture. We conclude that our selected exterior features improve the model performance. 

% Notably, the individual components of the F-LSTM did not outperform the single-modality methods proposed by \cite{mref1}. 
% The LSTM 3 had a prediction accuracy of 51\% compared to 53\%, and LSTM 1 had an accuracy of 68\% compared to 83\%. Yet, the F-LSTM significantly outperformed their fused model with an accuracy of 87.2\% compared to 75.5\% and an F1 score of 85.6\% compared to 73.2\%. 

The performance of the F-TF is expected due to the ability of the self-attention module to attend to long-range dependencies across each video sequence. \figref{transformervslstm} qualitatively supports this property. In the upper scenario, the model must understand the adjacent vehicle's relative speed to infer that it may still be on the driver's right despite not being visible in the road camera. In the bottom situation, the model discerns that the driver refrained from changing lanes when there was an opportunity in the past. This suggests that the driver is less likely to make a lane change in the future, especially when traffic is heavier. In both of these situations, the F-TF accurately forecasts that the driver will continue straight, a prediction that the F-LSTM fails to make correctly.

However, the F-TF has a high standard deviation across the validation splits and is not significantly better than our F-LSTM architecture. We believe this can be attributed to the lack of training data available in the Brains4Cars dataset. It is well-understood that computer vision transformer architectures~\cite{alexeyvit2020} require internet-scale amounts of data to significantly surpass convolutional neural network (CNN) architectures. We claim that because the transformer assumes no prior information about the sequential nature of the data, it requires far more data to learn this property and attain good performance. On the other hand, LSTM-based architectures are designed to leverage prior knowledge about the temporal relationship between frames, thus making it more data efficient. We postulate that our F-TF architecture would scale more effectively than other methods if provided with far more training data.

\begin{table}[htbp]
\caption{\label{tab:maintable} Zero time-to-maneuver accuracy and F1 score results.}
\renewcommand{\arraystretch}{1.2}
\begin{tabular}{p{1.6cm} c c c c c c}
\hline
Method &  Inside&  Outside&  Acc [\%]\ &  $\sigma$ &  F1 [\%]\ &  $\sigma$ \\
\hline
\multicolumn{7}{c}{Baseline Methods \cellcolor[gray]{0.9}} \\
\hline
Chance & - & - & 20 & - & 20 & -   \\
Prior  & - & - & 39 & - & - & -   \\
\hline
\multicolumn{7}{c}{Methods from \cite{brain4cars} and \cite{brain4cars2} \cellcolor[gray]{0.9}} \\
\hline
 IOHMM  & X & X & - & - &  72.7 & -   \\
 AIO-HMM  & X & X & - & - &  74.2 & -   \\
 S-RNN  & X & X & - & - &  74.4 & -   \\
 F-RNN-UL  & X & X & - & - &  78.9 & -   \\
 F-RNN-EL  & X & X & - & - &  80.6 & -   \\
\hline
\multicolumn{7}{c}{Methods from \cite{mref1} \cellcolor[gray]{0.9}} \\
\hline
 Outside  & - & X & 53.2 & 0.5 & 43.4 & 0.9   \\
 Inside  & X & - & 83.1 & 2.5 & 81.7 & 2.6   \\
 Two-stream  & X & X & 75.5 & 2.4 & 73.2 & 2.2   \\
\hline
\multicolumn{7}{c}{Method from \cite{mref3} \cellcolor[gray]{0.9}} \\
\hline
 Interior+VD  & X & - & 84.2 & - & 82.9 & -   \\
\hline
\multicolumn{7}{c}{Our Methods \cellcolor[gray]{0.9}} \\
\hline
 F-LSTM  & X & X & \textbf{87.2} & 2.3 & \textbf{85.6} & 3.4   \\
 F-TF & X & X & \textbf{87.5} & 4.9 & \textbf{86.3} & 4.5 \\
\end{tabular}
\renewcommand{\arraystretch}{1}
\end{table}

\subsection{Varying-time-to-maneuver}

\begin{figure}[htbp]
    \centering
    \includegraphics[width=\linewidth]{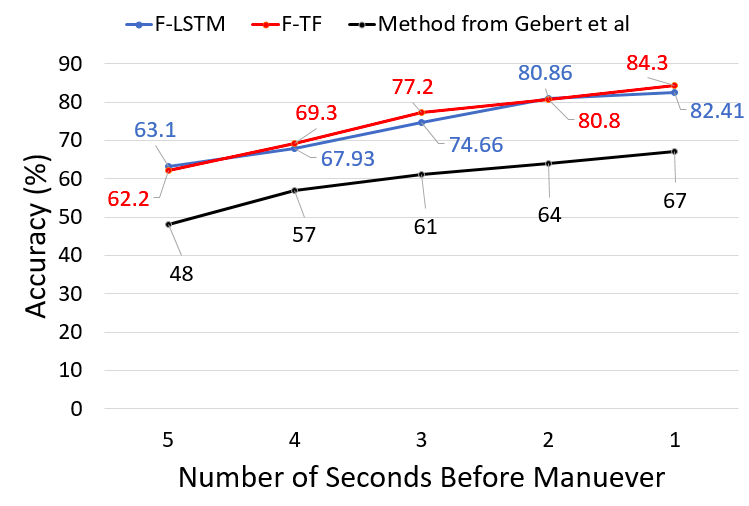}
    \caption{Comparison of our methods with state of the art for varying-time-to-maneuver}
    \label{fig:vttm}
\end{figure}

The results of the varying time-to-maneuver benchmark are shown below. Both the F-LSTM and the F-TF outperform the SOTA methods in this category. The F-LSTM and F-TF have an average prediction time of 4.34 and 4.35 seconds respectively compared to the 4.07 seconds from~\cite{mref1} and the 3.56 seconds from~\cite{mref3}. For a fair comparison with SOTA methods, we adopt their evaluation procedures. The accuracy of each method is compared using only the amount of data the method would have received at that time. At 5 seconds before the maneuver, each algorithm would have the first 30 frames of data of interior and exterior data. At 4 seconds before the maneuver, each algorithm would have access to the first 60 frames of data, and so on. The overall profile of the prediction accuracy over time is compared with the results from~\cite{mref1} and shown in Figure \ref{fig:vttm}.

These findings corroborate our hypothesis that sequential driver intention prediction benefits from having access to a good external feature representation. Compared to our method, which has a prediction accuracy of about 63\% 5 seconds before the maneuver takes place, we see that the prediction accuracy of the method proposed in \cite{mref1} is as low as 48\%. Our method outperforms the SOTA at every time interval. It is also worth noting that the decrease in performance when switching from a zero-time-to-maneuver problem to a varying-time-to-maneuver is much smaller for our proposed algorithms than the one proposed by \cite{mref1}. This decrease can be quantified by comparing the accuracy of the varying-time-to-maneuver model at 1 second with the zero-time model. The accuracy of the F-LSTM decreases by 5.5\% and the accuracy of the F-TF decreases by 3.7\%. The method proposed by \cite{mref1} degrades by 19.4\% in accuracy from 83.1\% to approximately 67.0\% between the zero-time-to-maneuver and varying-time-to-maneuver problem. This would suggest that our methods, particularly the F-TF, are more capable of forecasting driver intent.

%% file: conclusion.tex
\section{Conclusions}
In this work, we proposed a novel method to predict driver intentions across 5 driving maneuvers that fuses hand-crafted feature representations of the in-cabin and exterior cameras. Since driver intentions are, in general, difficult to predict, we show that prediction accuracy can be improved by incorporating multiple sources of information that the driver is likely considering instead of limiting the model to somatic information from the driver and dynamic information from the vehicle. We illustrate that our selection of external features complements the in-cabin features, which is different from previous methods that rely on learned exterior features. Our approaches substantially surpass the state-of-the-art methods in key metrics. The F-LSTM and F-TF architectures achieve an accuracy of 87.2\% and 87.5\% and are able to correctly predict the driver intention an average of 4.34 and 4.35 seconds before the maneuver occurs. This provides a key insight about what features are important for understanding driver intent. Interesting future directions to improve performance include developing better data augmentation strategies for additional data diversity, leveraging prior knowledge from pre-trained LSTM architectures to boost transformer learning efficiency, and expanding the driver action space.